\documentclass[runningheads]{llncs}

\usepackage[utf8]{inputenc}
\usepackage{enumerate}
\usepackage{algorithm}
\usepackage[noend]{algpseudocode}
\usepackage{varwidth}
\usepackage[utf8]{inputenc}

\usepackage{graphicx}
\begin{document}
\title{Is there Gender bias and stereotype in Portuguese Word Embeddings?}
%
%
\author{
Brenda Salenave Santana\inst{1}\orcidID{0000-0002-4853-5966} \and
Vinicius Woloszyn\inst{1}\orcidID{0000-0003-3554-5580}  \and
Leandro Krug Wives\inst{1}\orcidID{0000-0002-8391-446X}
}
\authorrunning{Santana et al.}
%
\institute{PPGC - Instituto de Informática - UFRGS, Porto Alegre RS, Brazil \\
\email{\{bssantana, vwoloszyn, wives\}@inf.ufrgs.br}\\
}
\maketitle              
\begin{abstract}
In this work, we propose an analysis of the presence of gender bias associated with professions in 
Portuguese word embeddings. The objective of this work is to study gender implications related to 
stereotyped professions for women and men in the context of the Portuguese language.
\end{abstract}
\section{Introduction}
Recently, the transformative potential of machine learning (ML) has propelled ML into the forefront of mainstream media. In Brazil, the use of such technique has been widely diffused gaining more space.
Thus, it is used to search for patterns, regularities or even concepts expressed in data sets \cite{goldschmidt2017data}, and can be applied as a form of aid in several areas of everyday life.

Among the different definitions, ML can be seen as the ability to improve performance in accomplishing a task through the experience \cite{mitchell1997machine}. Thus, \cite{carvalho2011inteligencia} presents this as a method of inferences of functions or hypotheses capable of solving a problem algorithmically from data representing instances of the problem. This is an important way to solve different types of problems that permeate computer science and other areas.

One of the main uses of ML is in text processing, where the analysis of the content the entry point for various learning algorithms. However, the use of this content can represent the insertion of different types of bias in training and may vary with the context worked. This work aims to analyze and remove gender stereotypes from word embedding in Portuguese, analogous to what was done in \cite{bolukbasi2016man} for the English language.

Hence, we propose to employ a public word2vec model pre-trained to analyze gender bias in the Portuguese language, quantifying biases present in the model so that it is possible to reduce the spreading of sexism of such models.
There is also a stage of bias reducing over the results obtained in the model, where it is sought to analyze the effects of the application of gender distinction reduction techniques.

This paper is organized as follows:
Section \ref{sec:relatedwork} discusses related works.
Section \ref{sec:embedding} presents the Portuguese word2vec embeddings model used in this paper and Section \ref{sec:proposal} proposes our method.
Section \ref{sec:results} presents experimental results, whose purpose is to verify results of a de-bias algorithm application in Portuguese embeddings word2vec model and a short discussion about it.
Section \ref{sec:finalremarks} brings our concluding remarks.



\section{Related Work} \label{sec:relatedwork}
There is a wide range of techniques that provide interesting results in the context of ML algorithms geared to the classification of data without discrimination; these techniques range from the pre-processing of data \cite{Kamiran2012} to the use of bias removal techniques\cite{4909197} in fact.
Approaches linked to the data pre-processing step usually consist of methods based on improving the quality of the dataset after which the usual classification tools can be used to train a classifier.
So, it starts from a baseline already stipulated by the execution of itself. On the other side of the spectrum, there are Unsupervised and semi-supervised learning techniques, that are attractive because they do not imply the cost of corpus annotation \cite{woloszyn2018distrustrank,dos2018ddc,woloszyn2017mrr,woloszyn2017beatnik}.


The bias reduction is studied as a way to reduce discrimination through classification through different approaches \cite{pedreshi2008discrimination} \cite{calders2010three}.
In \cite{DBLP:journals/corr/abs-1710-06921} the authors propose to specify, implement, and evaluate the ``fairness-aware" ML interface called themis-ml.
In this interface, the main idea is to pick up a data set from a modified dataset.
Themis-ml implements two methods for training fairness-aware models.
The tool relies on two methods to make agnostic model type predictions: Reject Option Classification and Discrimination-Aware Ensemble Classification, these procedures being used to post-process predictions in a way that reduces potentially discriminatory predictions.
According to the authors, it is possible to perceive the potential use of the method as a means of reducing bias in the use of ML algorithms.

In \cite{bolukbasi2016man}, the authors propose a method to hardly reduce bias in English word embeddings collected from Google News.
Using word2vec, they performed a geometric analysis of gender direction of the bias contained in the data.
Using this property with the generation of gender-neutral analogies, a methodology was provided for modifying an embedding to remove gender stereotypes.
Some metrics were defined to quantify both direct and indirect gender biases in embeddings and to develop algorithms to reduce bias in some embedding.
Hence, the authors show that embeddings can be used in applications without amplifying gender bias.

\section{Portuguese Embedding} \label{sec:embedding}
In \cite{mikolov2013efficient}, the quality of the representation of words through vectors in several models is discussed. According to the authors, the ability to train high-quality models using simplified architectures is useful in models composed of predictive methods that try to predict neighboring words with one or more context words, such as Word2Vec.
Word embeddings have been used to provide meaningful representations for words in an efficient way.

In \cite{DBLP:journals/corr/abs-1708-06025}, several word embedding models trained in a large Portuguese corpus are evaluated. Within the Word2Vec model, two training strategies were used. In the first, namely Skip-Gram, the model is given the word and attempts to predict its neighboring words. The second, Continuous Bag-of-Words (CBOW), the model is given the sequence of words without the middle one and attempts to predict this omitted word. The latter was chosen for application in the present proposal.

The authors of \cite{DBLP:journals/corr/abs-1708-06025} claim to have collected a large corpus from several sources to obtain a multi-genre corpus representative of the Portuguese language.
Hence, it comprehensively covers different expressions of the language, making it possible to analyze gender bias and stereotype in Portuguese word embeddings.
The dataset used was tokenized and normalized by the authors to reduce the corpus vocabulary size, under the premise that vocabulary reduction provides more representative vectors.

\section{Proposed Approach} \label{sec:proposal}

Some linguists point out that the female gender is, in Portuguese, a particularization of the masculine.
In this way the only gender mark is the feminine, the others being considered without gender (including names considered masculine).
In \cite{hellinger2015gender} the gender representation in Portuguese is associated with a set of phenomena, not only from a linguistic perspective but also from a socio-cultural perspective. Since most of the termination of words (\textit{e.g.}, advogad\underline{a} and advogad\underline{o}) are used to indicate to whom the expression refers, stereotypes can be explained through communication.
This implies the presence of biases when dealing with terms such as those referring to professions.

\begin{figure}[htb]
\centering
\includegraphics[scale=0.45]{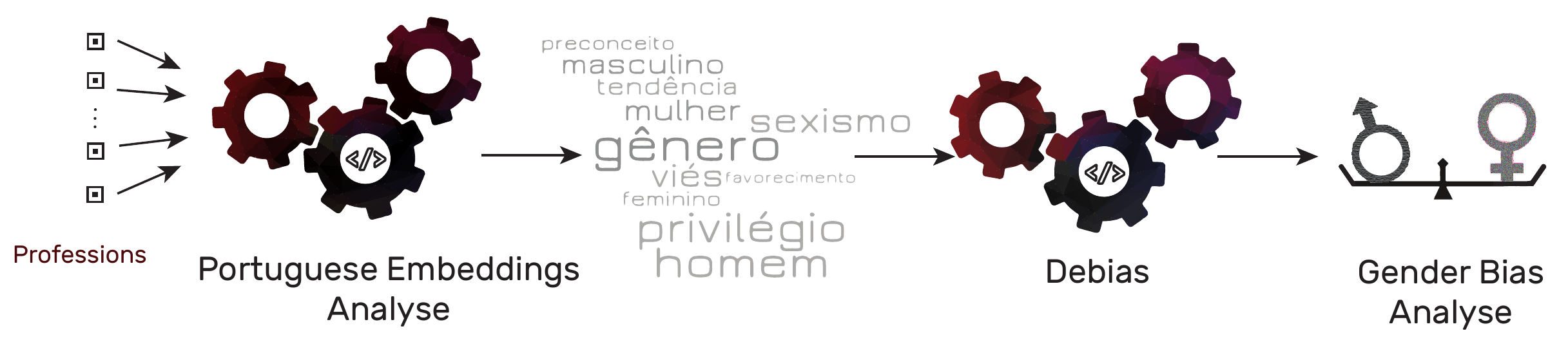}
\caption{Proposal}
\label{fig:proposal}
\end{figure}

Figure \ref{fig:proposal} illustrates the approach proposed in this work. 
First, using a list of professions relating the identification of female and male who perform it as a parameter, we evaluate the accuracy of similarity generated by the embeddings. 
Then, getting the biased results, we apply the De-bias algorithm \cite{bolukbasi2016man} aiming to reduce sexist analogies previous generated.
Thus, all the results are analyzed by comparing the accuracies.

Using the word2vec model available in a public repository \cite{DBLP:journals/corr/abs-1708-06025}, the proposal involves the analysis of the most similar analogies generated before and after the application of the \cite{bolukbasi2016man}.
The work is focused on the analysis of gender bias associated with professions in word embeddings.
So therefore into the evaluation of the accuracy of the associations generated, aiming at achieving results as good as possible without prejudicing the evaluation metrics.


Algorithm \ref{alg} describes the method performed during the evaluation of the gender bias presence.
In this method we try to evaluate the accuracy of the analogies generated through the model, that is, to verify the cases of association matching generated between the words.

 \begin{algorithm}[!htb]
\caption{Model Evaluation}\label{alg}
\begin{algorithmic}[1]

\Function{w2v\_evaluate}{$model, professions\_pair$}
  \State open\_model($model$)
  \State count = $0$
  \For{$female, male$ in $profession\_pairs$}\Comment{read list of tuples}
  \State x = model.most\_similar(positive=[`ela', male], negative=[`ele']) 

    \If{x = female}
      \State count += 1 
    \EndIf
  \EndFor
  \State accuracy = count/size(profession\_pairs)
  \State \textbf{return} accuracy
  
  \EndFunction
\end{algorithmic}
\label{algorithm}
\end{algorithm}



  

\section{Experiments} \label{sec:results}
The purpose of this section is to perform different analysis concerning bias in word2vec models with Portuguese embeddings.
The Continuous Bag-of-Words model used was provided by \cite{DBLP:journals/corr/abs-1708-06025} (described in Section \ref{sec:embedding}).
For these experiments, we use a model containing $934966$ words of dimension $300$ per vector representation.
To realize the experiments, a list containing fifty professions labels for female and male was used as the parameter of similarity comparison.

Using the python library gensim\footnote{Available in: https://pypi.org/project/gensim/}, we evaluate the extreme analogies generated when comparing vectors like:
$\overrightarrow{mulher} - \overrightarrow{x} \approx \overrightarrow{y} - \overrightarrow{homem}$, 
where $x$ represents the item from professions list and $y$ the expected association.
The most similarity function finds the top-N most similar entities, computing cosine similarity between a simple mean of the projection weight vectors of the given docs.
Figure \ref{fig:extreme_analogies} presents the most extreme analogies results obtained from the model using these comparisons.

Applying the Algorithm \ref{alg}, we check the accuracy obtained with the similarity function before and after the application of the de-bias method.
Table \ref{tab:accuracy} presents the corresponding results.
In cases like the analogy of `garçonete' to `stripper' (Figure \ref{fig:extreme_analogies}, line 8), it is possible to observe that the relationship stipulated between terms with sexual connotation and females is closer than between females and professions.
While in the male model, even in cases of non-compliance, the closest analogy remains in the professional environment.

\begin{table}[htb]
\centering
\caption{Accuracy Obtained in Predicting Model Analogies}
\begin{tabular}{
c|c} \hline
\textbf{Model}  &  \multicolumn{1}{|c}{\textbf{Accuracy}} \\ \hline
Before Debias & 18.18 \% \\
After Debias  & 03.03 \% \\
\hline
\end{tabular}
\label{tab:accuracy}
\end{table}

Using a confidence factor of 99\%, when comparing the correctness levels of the model with and without the reduction of bias, the prediction of the model with bias is significantly better.
Different authors \cite{kamishima2012fairness}\cite{zliobaite2015survey} show that the removal of bias in models produces a negative impact on the quality of the model.
On the other hand, it is observed that even with a better hit rate the correctness rate in the prediction of related terms is still low.


\begin{figure}
\centering
\begin{varwidth}[t]{.5\textwidth}
\textbf{Extreme `he'}
\begin{enumerate}
\item escritor $\rightarrow$ poeta
\item cantor$\rightarrow$  músico
\item pintor $\rightarrow$ escultor  
\item secretario $\rightarrow$ secretário  
\item ator $\rightarrow$ actor  
\item historiador $\rightarrow$ poeta  
\item arquiteto$\rightarrow$ arquitect 
\item fotógrafo $\rightarrow$ cineasta 
\item advogado $\rightarrow$ empresário 
\item juiz $\rightarrow$ juíz  
\end{enumerate}

\end{varwidth}
\hspace{4em}
\begin{varwidth}[t]{.5\textwidth}

\textbf{Extreme `she'}
\begin{enumerate}
\item atriz $\rightarrow$ actriz
\item escritora $\rightarrow$ poetisa
\item pesquisadora $\rightarrow$ bióloga
\item sindica $\rightarrow$ medicamen
\item diretora $\rightarrow$ coordenadora
\item matemática $\rightarrow$ astronomia
\item historiadora $\rightarrow$ pesquisadora
\item garçonete $\rightarrow$ stripper
\item secretaria $\rightarrow$ secretária
\item enfermeira $\rightarrow$ psicóloga
\end{enumerate}
\end{varwidth}
\caption{Extreme Analogies}
\label{fig:extreme_analogies}
\end{figure}

\section{Final Remarks} \label{sec:finalremarks}
This paper presents an analysis of the presence of gender bias in Portuguese word embeddings.
Even though it is a work in progress, the proposal showed promising results in analyzing predicting models.

A possible extension of the work involves deepening the analysis of the results obtained, seeking to achieve higher accuracy rates and fairer models to be used in machine learning techniques.
Thus, these studies can involve tests with different methods of pre-processing the data to the use of different models, as well as other factors that may influence the results generated.
This deepening is necessary since the model's accuracy is not high.

To conclude, we believe that the presence of gender bias and stereotypes in the Portuguese language is found in different spheres of language, and it is important to study ways of mitigating different types of discrimination.
As such, it can be easily applied to analyze racists bias into the language, such as different types of preconceptions.

%
%
%
 \newpage
 \nocite{*}
 \bibliographystyle{splncs04}
 \bibliography{bibliography}
\end{document}